\documentclass[11pt,a4paper]{article}
\usepackage[hyperref]{naaclhlt2018}
\usepackage{times}
\usepackage{latexsym}

\usepackage{url}
\usepackage{booktabs}
\usepackage{graphicx}
\graphicspath{ {./img/} }
\usepackage[linesnumbered,ruled]{algorithm2e}
\usepackage{amsmath}

\aclfinalcopy % Uncomment this line for the final submission
\title{Using Inter-Sentence Diverse Beam Search to Reduce Redundancy in Visual Storytelling}

\author{Chao-Chun Hsu, Szu-Min Chen, Ming-Hsun Hsieh, Lun-Wei Ku\\
  Academia Sinica, Taiwan \\
   \{joe32140, b02902026, troutman, lwku\}@iis.sinica.edu.tw}

\date{}

\begin{document}
\maketitle

\begin{abstract}
  Visual storytelling includes two important parts: 
  coherence between the story and images as well as the story structure.
  For image to text neural network models, 
  similar images in the sequence would provide close information for story generator to obtain almost identical sentence.
  However, repeatedly narrating same objects or events will undermine a good story structure. 
  In this paper, we proposed an inter-sentence diverse beam search to generate a more expressive story.
  Comparing to some recent models of visual storytelling task, 
  which generate story without considering the generated sentence of the previous picture, 
  our proposed method can avoid generating identical sentence even given a sequence of similar pictures.

\end{abstract}

\section{Introduction}

Visual storytelling is gaining more popularity in recent years.
The task requires machines to both comprehend the content of a stream of images and also produce 
a narrative story. The recent rapid progress of the neural networks has enabled the models to achieve 
promising performance on the task of image captioning \cite{xu2015show,vinyals2017show}, which is 
also an image-to-text problem. Nonetheless, visual storytelling is much more difficult and complicated; 
besides generating an individual sentence solely, to form a complete story needs to take the coherency as well as 
the main focus of the story and even the creativity into consideration.

Park and Kim\cite{park2015expressing} viewed this problem 
as a retrieval task and incorporating the information of discourse entities to model the coherence. Some researchers 
designed a variation of GRU which can "skip" some input which strengthens the ability to deal with longer dependency\cite{liu2016storytelling}.
However, retrieval-based methods are less general and limited to those seen sentences. In a task where the output varies dramatically,
using a more flexible manner to generate stories seems to be more appropriate. Huang\cite{huang2016visual} published a large 
dataset containing photo streams where each of them is paired with a story. They also proposed a neural baseline model of this task which based on the
seq-to-seq framework. First encoding all images, and then they use the encoder hidden state as initial hidden state of decoder 
GRU and produce the whole story. In addition to visual grounding, 
we expect the generated stories can be as similar as possible to those written by human. This may be breakdown to several characteristics
such as the style, the repeated use of words, or how detailed the paragraph is, etc. It is hard to enumerate all possible features, 
thus, the work \cite{wang2018show} utilize the Seq-Gan framework, designing two discriminators, one is responsible for the degree of matching
of an image and a sentence, the other focuses on the text only, trying to mimic the language style of human. More recent work analyzing
the relation between scores of automatic metrics such as BLEU \cite{papineni2002bleu} and those from human evaluation. 
They utilize Inverse Reinforcement Learning framework, attempting to "learn" a reward by adversarial network which is similar to the criterion of human judgment \cite{wang2018no}.

However, these neural models aims to learn a human distribution and have nothing to do with the creativity and also
consumes lots of computing resource and time. In this paper,
we proposed an inter-sentence diverse beam search which can generate interesting sentences and avoid redundancy given a sequence of similar photos.
Comparing to the diverse beam search which generates different groups of sentence given a condition, 
our inter-sentence version focuses on producing various sub-stories given a sequence of images.
The proposed model improves the max meteor scores on visual storytelling dataset from 29.3\% to 31.7\% comparing to the baseline model.

\begin{table}[]
\centering
\resizebox{\columnwidth}{!}{%
\begin{tabular}{llllll}

 & \includegraphics[width=0.21\textwidth]{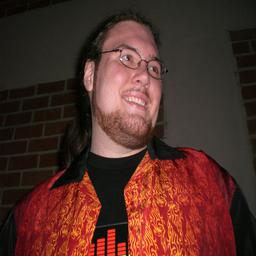} 
 & \includegraphics[width=0.21\textwidth]{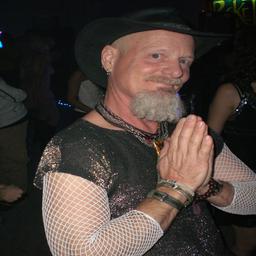}
 & \includegraphics[width=0.21\textwidth]{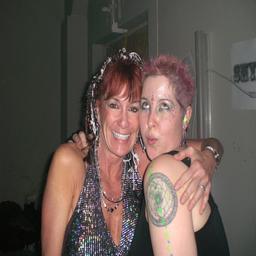}
 & \includegraphics[width=0.21\textwidth]{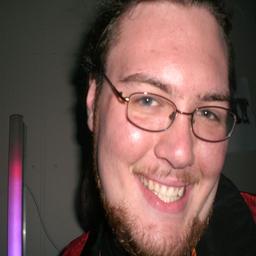}
 & \includegraphics[width=0.21\textwidth]{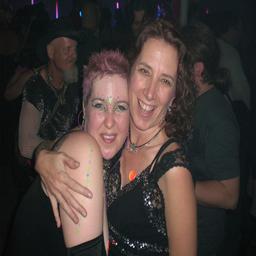}\\ \hline 

\textbf{baseline} & \begin{tabular}[c]{@{}l@{}}the friends were excited\\ to go out for a night .\end{tabular} & \begin{tabular}[c]{@{}l@{}}we had a lot of fun .\end{tabular} & \begin{tabular}[c]{@{}l@{}}we had a great time .\end{tabular} & \begin{tabular}[c]{@{}l@{}} we had a great time .\end{tabular} & \begin{tabular}[c]{@{}l@{}}we all had a great time .\end{tabular} \\ \hline
\end{tabular}%
}
\caption{Repetitive story generated by baseline model}
\label{table:exmaple_story}
\end{table}

\section{Method}

\subsection{Basic Architecture}
\begin{figure} []
\centering
\includegraphics[width=1.0\linewidth]{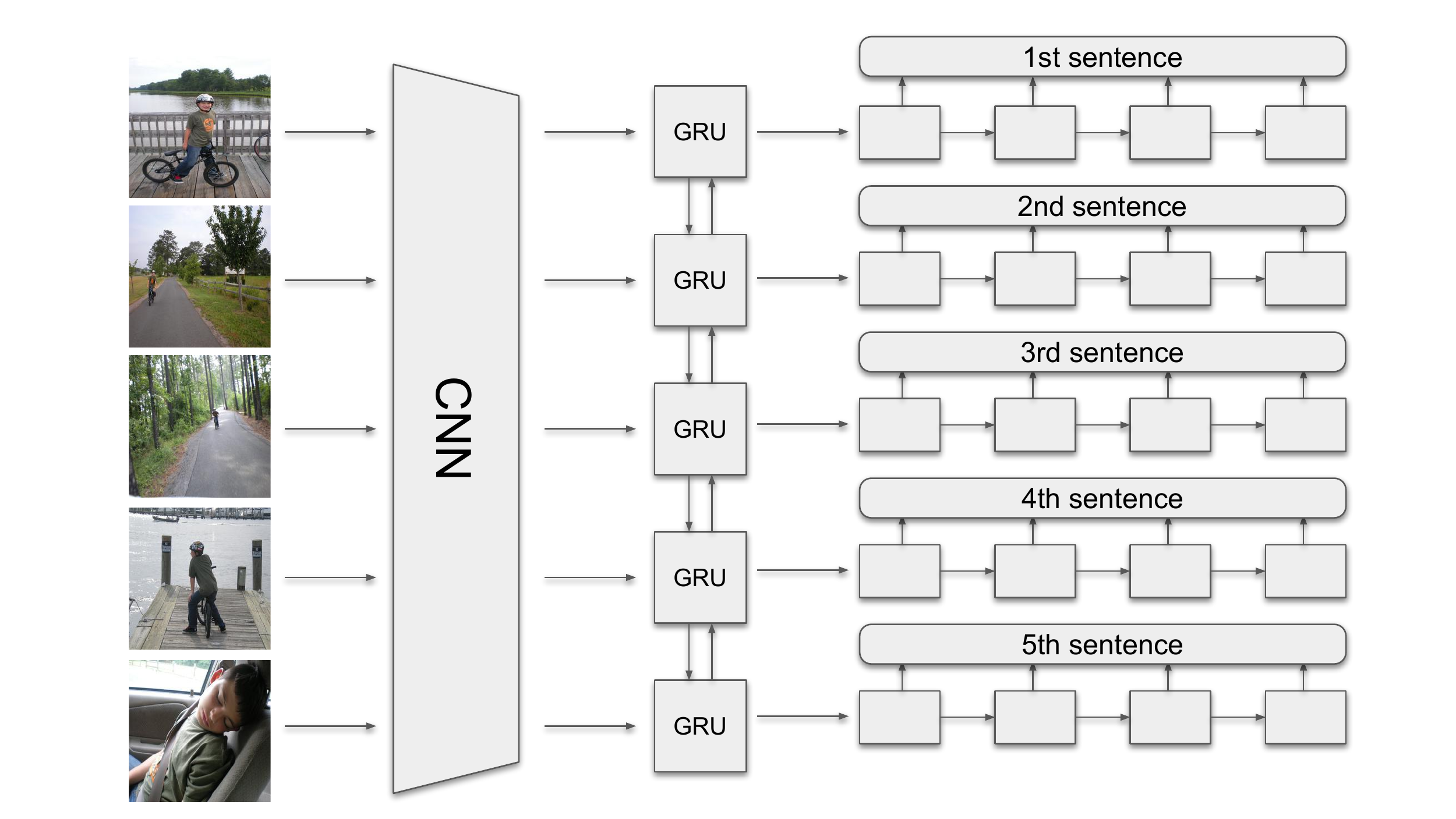}
\caption{The architecture of baseline model}
\label{fig:our_awesome_model}
\end{figure}

The baseline model is a encoder-decoder framework as in Figure \ref{fig:our_awesome_model}. 
We then apply proposed method on the top of the baseline model.

\paragraph{Encoder}
We utilize the pretrained Resnet-152 and extract the output of the second-to-last fully connected layer
to form an 2048-dimensional image representation vector. Since a story is composed of five images, the model should take the order of them into 
consideration as well as memorize the content of previous pictures. Thus, a bidirectional GRU will take as input the five image vectors and produce
five context-aware image embeddings.
\paragraph{Decoder}
Given an image embedding, another GRU is used to produce one sentence at a time. 
To form a complete story, 
we first generate five sub-stories separately and then concatenating them. 
We have tried both using image embedding as the initial hidden state and concatenated the image embedding with the word embedding 
as the input of each time steps. 
The latter one yields a smoother validation curve and thus we adopted this setting in our last submission.

\subsection{Decoding techniques}
We have noticed a major issue of the baseline model is that the generated sentences may repeat themselves as in Table~\ref{table:exmaple_story}
because of the similar image or generic story. E.g. \textit{I have a great time}.
To overcome this, we proposed a variation of beam search which takes into account the previous generated sentences. 

Inspired by the diverse beam search method \cite{vijayakumar2016diverse}, our model will aware of the words
used in previous incomplete story when decoding one sub-story, and then calculating the score based on each next word's probability and the diversity penalty, which will discuss below.
Afterward, the model will rearrange the candidates according to the scores and selects the proper word.
In our setting, we use bag-of-word to represent the previous sentences and adopting the relatively simple
\textit{hamming diversity}, which punishes a selected token proportional to its previous occurrences,
but any kind of penalty can be plugged into this framework.
Different from \shortcite{vijayakumar2016diverse}, their work deals with intra-sentences diversity while ours focus on the inter-sentences diversity, which
is more suitable for this task. 

\paragraph{Inter-sentence Diverse Beam Search} For the decoding of the first sentence, we perform a regular beam search with no diversity penalty. 
After that, we consider the diversity with all previous sentences for the following sentence generation process.

From a finite vocabulary $\mathcal{V}$ and an input $x$, the model output the sequence y based on the probability P(y$|$x) where the output sequence y = $(y_1,...,y_T)$. 
Let $\theta(y_\gamma)=log Pr (y_t|x,y_1,...,y_{[t-1]})$, we denote $\Theta(\text{y}_{[t]}) = \Sigma_{\gamma\in[t]}\theta(y_\gamma)$.
At each time step $t$, the set $Y_{t}^i = \{y_{1,[t]}^i,...,y_{B_,[t]}^i\}$ is updated by considering the set ${\mathcal{Y}}_{t}^{i}=Y_{t}^i \times \mathcal{V} $ for the image $i$ with $B$ beams. Diverse beam search adds diversity penalty by calculating sentences with diversity function $\Delta$ multiplied by the diversity strength $\lambda$.
Using the notation of \cite{vijayakumar2016diverse}, each time step $t$ of this process for image $i$ can be presented as,

\begin{algorithm*}
\small
\SetAlgoLined
 \caption{Inter-Sentence Diverse Beam Search}
 Perform inter-sentences diverse beam search with $I$ images using beam width $B$\\
 \label{alg:math_here_lol}
 \tcp{beam search without diversity penalty for the first image}
 $Y_{[T]}^1 \leftarrow BeamSearch(x^1)$\\
  \tcp{Perform diverse beam search for the following images} 
 \For{i=2,...I}{
  \For{t=1,...T}{
   $\Theta(\text{y}_{b,[t]}^i) \leftarrow \Theta(\text{y}_{b,[t]}^i) + \lambda \Delta(Y_{[T]}^1,...Y_{[T]}^{i-1})[y_{b,t}^i]
   \quad b\in[B],y_{b,[t]}^i\in {\mathcal{Y}}_{t}^{i} \quad \text{and} \quad \lambda > 0$
   
   $Y_{[t]}^i \leftarrow \operatorname*{arg\,max}_{(y_{1,[t]}^i,...y_{B,[t]}^i)}\sum_{b\in[B]} \Theta(\text{y}_{b,[t]}^i)$
  }
 }
 Return set of $I$ sentences $Y_{[T]} = \bigcup_{i=1}^I Y_{[T]}^i $\\
\end{algorithm*}

\begin{table*}[]
\centering
\resizebox{\textwidth}{!}{%
\begin{tabular}{llllll}
\toprule
 & \includegraphics[width=0.21\textwidth]{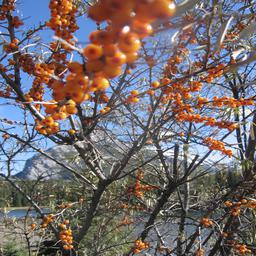}
 & \includegraphics[width=0.21\textwidth]{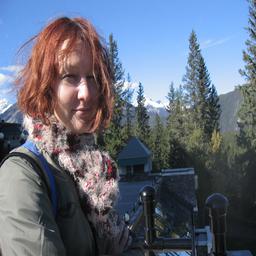}
 & \includegraphics[width=0.21\textwidth]{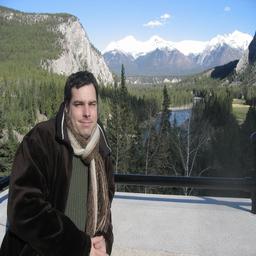}
 & \includegraphics[width=0.21\textwidth]{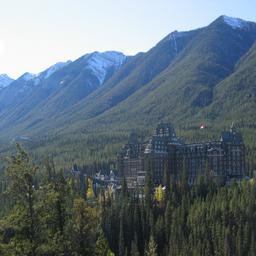}
 & \includegraphics[width=0.21\textwidth]{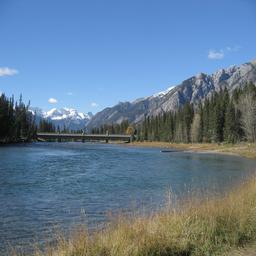}\\ \hline 
baseline & the trees were beautiful. & \begin{tabular}[c]{@{}l@{}}we took a picture of \\ the mountains.\end{tabular} & \begin{tabular}[c]{@{}l@{}}we took a picture of \\the mountains.\end{tabular} & \begin{tabular}[c]{@{}l@{}}the mountains were \\beautiful and the.\end{tabular} & \begin{tabular}[c]{@{}l@{}}the river was \\the lake  was beautiful.\end{tabular} \\ \hline
\textbf{ours} & i went on a hike last week. & \begin{tabular}[c]{@{}l@{}}we had to take a picture \\of the mountains.\end{tabular} & \begin{tabular}[c]{@{}l@{}}they took pictures of the\\ lake and scenery.\end{tabular} & \begin{tabular}[c]{@{}l@{}}this is my favorite part of the \\trip with my wife ,i 've never \\ seen such a beautiful view! \end{tabular} & \begin{tabular}[c]{@{}l@{}}it was very peaceful \\and serene.\end{tabular} \\ \hline
\end{tabular}%
}
\caption{Generated stories from two models.}
\label{table:exmaple_compare_story}
\end{table*}

% the fucking equation
\begin{equation}
\begin{aligned}
Y_{[t]}^i \leftarrow \operatorname*{arg\,max}_{(y_{1,[t]}^i,...y_{b,[t]}^i)\in {\mathcal{Y}}_{t}^{i}} \sum_{b\in[B]} \Theta(\text{y}_{b,[t]}^i) \\
+ \lambda \Delta(Y_{[T]}^1,...Y_{[T]}^{i-1})[y_{b,t}^i] \\ 
\quad \lambda > 0
\end{aligned}
\end{equation}
In our experiments, we applied hamming diversity for $\Delta$ and found $\lambda$ achieved better results at 2.
Our purposed inter-sentence diverse beam search is illustrated in Algorithm~\ref{alg:math_here_lol}.
% Please add the following required packages to your document preamble:
% \usepackage{graphicx}

\section{Visual Storytelling Dataset}
Visual Storytelling dataset(VIST) is the first dataset of image sequence paired with text:
(1) Descriptions of imagesin-isolation (DII); 
and (2) Stories for images-insequence (SIS).
The dataset contain more than 20,000 unique photos in 50,000 sequences.
We eliminate the stories that have broken images($\sim1\%$).

\section{Experiments and results}
\subsection{Model Setup}
We first scale the images to 224*224 and apply horizontal flip into training process. 
Then the images is normalized to fit in pretrained resnet-152 CNN model.
For the hyperparameters, the image feature size is 256, and the hidden size of decoder GRU cell is 512.
The word appearing more than three times in the corpus is pick into vocabulary.
We select Adam as our optimizer and set learning rate to 2e-4. Schedule sampling and batch normalization are introduced in the training process.

\subsection{Result}
Every photo sequence in test set has 2 to 5 reference stories. 
We evaluate our models by max meteor score of references for a photo sequence.
As we can see in Table~\ref{table:max_meteor}, the inter-sentence diverse beam search improve the max meteor score from 29.3 to 31.7.
Besides the improvement on metric, the proposed method can generate interesting sentences (Table~\ref{table:exmaple_compare_story}).

\begin{table}[]
\small
\centering
%\resizebox{\textwidth}{!}{%
\begin{tabular}{|l|l|l|}
\hline
 & Baseline & Ours \\ \hline
Meteor & 29.3 & 31.7 \\ \hline
\end{tabular}%
%}
\caption{Max meteor score(\%)}
\label{table:max_meteor}
\end{table}

\section{Conclusion and Future Work}
We proposed a new decoding approach for the visual storytelling task, which avoided to generate the repeated information of the photo sequence. Instead, our model would attempt to produce a diverse expression for the image.

Nevertheless, the value of diversity weight $\lambda$ requires human heuristics to determine the trade-off between the diversity and the output sequence probability. Future efforts should be dedicated to introduce a data-driven method to make machines learn to put the attention on either the sequence probability or the distinct details in the image.

\bibliography{naaclhlt2018}
\bibliographystyle{acl_natbib}

\end{document}